\documentclass{article}

\usepackage[preprint]{neurips_2023}
\usepackage{booktabs}
\usepackage{graphicx}
\usepackage{float}
\usepackage{subfigure}
\usepackage{wrapfig}
\usepackage{makecell}

\usepackage{hyperref}

\usepackage[utf8]{inputenc} 
\usepackage[T1]{fontenc}    
\usepackage{url}            
\usepackage{booktabs}       
\usepackage{amsfonts}       
\usepackage{nicefrac}       
\usepackage{microtype}      
\usepackage{xcolor}         
\usepackage{amsmath}
\usepackage{multirow}
\usepackage{amsfonts,amssymb}
\usepackage{booktabs}
\usepackage{multirow}
\usepackage{graphicx}

\usepackage{natbib}
\setcitestyle{numbers,square,comma}

\usepackage{color}
\definecolor{lightblue}{RGB}{0, 110, 204}
\definecolor{blue}{RGB}{0, 0, 255}
\hypersetup{colorlinks=true,linkcolor=blue, linktocpage}

\newcommand{\kits}{{KiTS21}}
\newcommand{\msd}{{MSD-Pancreas}}

\usepackage{hyperref}

\hypersetup{
    colorlinks=true,
    linkcolor=blue,
    filecolor=magenta,      
    urlcolor=blue,
    pdftitle={Overleaf Example},
    pdfpagemode=FullScreen,
    }

\title{3DSAM-adapter: Holistic Adaptation of SAM from \\ 2D to 3D for Promptable Medical Image Segmentation}

\author{%
  Shizhan Gong, Yuan Zhong, Wenao Ma, Jinpeng Li, Zhao Wang, \\ \textbf{Jingyang Zhang, Pheng-Ann Heng, Qi Dou} \\
  Department of Computer Science and Engineering\\
  The Chinese University of Hong Kong\\
}

\begin{document}

\maketitle

\begin{abstract}

Despite that the segment anything model (SAM) achieved impressive results on general-purpose semantic segmentation with strong generalization ability on daily images, its demonstrated performance on medical image segmentation is less precise and not stable, especially when dealing with tumor segmentation tasks that involve objects of small sizes, irregular shapes, and low contrast. Notably, the original SAM architecture is designed for 2D natural images, therefore would not be able to extract the 3D spatial information from volumetric medical data effectively. In this paper, we propose a novel adaptation method for transferring SAM from 2D to 3D for promptable medical image segmentation. Through a holistically designed scheme for architecture modification, we transfer the SAM to support volumetric inputs while retaining the majority of its pre-trained parameters for reuse. The fine-tuning process is conducted in a parameter-efficient manner,  wherein most of the pre-trained parameters remain frozen, and only a few lightweight spatial adapters are introduced and tuned. Regardless of the domain gap between natural and medical data and the disparity in the spatial arrangement between 2D and 3D, the transformer trained on natural images can effectively capture the spatial patterns present in volumetric medical images with only lightweight adaptations. We conduct experiments on four open-source tumor segmentation datasets, and with a single click prompt, our model can outperform domain state-of-the-art medical image segmentation models on 3 out of 4 tasks, specifically by 8.25\%, 29.87\%, and 10.11\% for kidney tumor, pancreas tumor, colon cancer segmentation, and achieve similar performance for liver tumor segmentation. We also compare our adaptation method with existing popular adapters, and observed significant performance improvement on most datasets.
Our code and models are available at: \url{https://github.com/med-air/3DSAM-adapter}. 

\end{abstract}

\section{Introduction}

Foundation models trained on massive data have demonstrated impressive ability on various general tasks~\cite{openai2023gpt4,ramesh2022hierarchical,wang2023detecting}, and are envisaged to impact downstream domains, especially where data collection and labeling are expensive.
Segment anything model (SAM)~\cite{kirillov2023segany} is the one from the computer vision field, which has shown success for general-purpose promptable object segmentation.
As is known, such powerful discrimination ability relies on the coverage of distributions as exhibited in training data. This is also the underlying reason for the reported suboptimal performance when applying SAM to domain-specific tasks such as medical image segmentation~\cite{mazurowski2023segment,ma2023segment,huang2023segment,he2023accuracy,zhang2023segment,ji2023segment,cheng2023sam}. 
For instance, Huang et al.~\cite{huang2023segment} extensively tested SAM on 52 medical datasets and observed limited performance on objects with irregular shapes or limited contrast. Mazurowski et al.~\cite{mazurowski2023segment} evaluated SAM under different prompt settings, and showed unstable results upon single-point prompts on 3D medical images.
Therefore, adaptation for out-of-distribution domains is needed, but how to design the adapter is yet unclear.



The success of SAM can be partly ascribed to the powerful prompt engineering, which originates from large language models~\cite{liu2023pre} with recent variants improving model generalizability across tasks.
SAM uses the form of positional prompts manifested in a point click or bounding box, so that the model predicts a segmentation mask for where the prompt is provided. The problem of such positional prompts is the lack of high-level semantic awareness, because the segmentor tends to discriminate nearby structures relying on local features such as edges, shapes, or contrast~\cite{sakinis2019interactive}.
This is the case for SAM pre-training data where the objects mostly have clear boundaries, but are not suitable for medical images because the boundaries between the tumors and their surrounding tissues are often ambiguous. 
Therefore, adaptation or model redesign is required to transfer SAM to specific applications with domain knowledge considered.
Recent works on parameter-efficient adaptation methods have tried to update pre-trained parameters via learning task-specific vision prompts~\cite{nie2022pro}, specifying a small proportion of parameters to tune~\cite{zaken2021bitfit}, or incorporating lightweight plug-and-play adapters~\cite{houlsby2019parameter,NEURIPS2022_ee604e1b}. Promising results have been achieved even though just a small amount of parameters are fine-tuned or added, which motivates us to also seek an efficient version of \textbf{3DSAM-adapter} for medical imaging.

How to adapt SAM from 2D to 3D for medical image segmentation should be carefully considered from all aspects in order to fit volumetric data. Related work includes methods adapting foundation models from 2D images to 3D videos~\cite{yang2023aim}, for example, Pan et al.~\cite{pan2022st} modify a standard adapter by incorporating depth-wise convolutions to impose spatial-temporal reasoning.
Domain experts on medical imaging also proposed 2D to 3D adapters, for example, Want et al.~\cite{wang2023med} develop a 3D convolution-based adapter together with Fast Fourier Transform to extract spatial information. Wu et al.~\cite{wu2023medical} replicate the weights learned from 2D images to help fuse spatial information of the third dimension.
However, their main backbones (especially transformer-based) are still based on 2D, whereas the 3D information is compensated through additional fusion structures. This may work for video data where the temporal dimension is essentially different from spatial dimensions, but is definitely sub-optimal for medical images where the 3-dimensional spatial information is isotropic. 
To date, how to effectively adapt parameters pre-trained on 2D images to catch 3D spatial information is not explored, due to challenging requirements on holistic modification of the network which would affect many pre-trained weights. In addition, raising the dimension can greatly increase the number of tokens in transformation blocks which also potentially leads to numerical and memory issues.


In this paper, we propose a new parameter-efficient adaptation method to holistically adapt SAM from 2D to 3D for medical image segmentation. First, for the image encoder at the input level, we precisely design the modification scheme to allow the original 2D transformer supports volumetric inputs, while keeping as many as possible pre-trained weights reusable. We find that
weights pre-trained on 2D images can still capture some 3D spatial patterns through parameter-efficient fine-tuning. Second, for the prompt encoder level, instead of using positional encoding as the prompt representation, we propose a visual sampler from the image embedding to serve as the representation of the point prompt, and further use a set of global queries to eliminate noisy prompts. This strategy proves to behave well to overcome the over-smoothing issues caused by a drastic increase in image token size accompanying the dimension raising, and also improves the model's robustness to inaccurate prompts. Last, for the mask decoder at the output level, we emphasize a lightweight design, with the promotion of adding multi-layer aggregation. We conduct experiments on medical tumor segmentation datasets with comprehensive comparisons with domain SOTA approaches including nn-UNet~\cite{isensee2021nnu}, as well as recent adapters in general. The results show our methods can outperform existing methods by a large margin. The method also shows robustness on the number and position of the prompt. For instance, a single point at the margin of the tumor can also serve as a prompt for accurate segmentation.
Our major contributions are summarized as: 
\begin{itemize}
\item We propose a holistic 2D to 3D adaptation method via carefully designed modification of  SAM architecture, which adds only 7.79\% more parameters and keeps most of the pre-trained weights reusable while performing well for volumetric medical image segmentation.
\item We introduce a novel parameter-efficient fine-tuning method to effectively capitalize a large image model pre-trained on 2D images for 3D medical image segmentation with only 16.96\% tunable parameters (including newly added parameters) of the original model.
\item We conducted experiments with four datasets for medical image segmentation. Results show that our 3DSAM-adapter significantly outperforms nn-UNet~\cite{isensee2021nnu} on three of the four datasets (by 8.25\% for kidney tumor, 29.87\% for pancreas tumor and 10.11\% for colon cancer) and comparable on the liver tumor. We also demonstrate superior performance of our proposed method over recent parameter-efficient fine-tuning methods such as ST-adapter~\cite{pan2022st}.
\end{itemize}


\section{Related Works}

\noindent \textbf{Foundation models in computer vision.}
With the breakthroughs in deep learning models, most modern vision models follow the pre-training and fine-tuning paradigm~\cite{he2019rethinking,hulora}. Large and generalizable foundation models have seen significant interest in computer vision, benefit from pre-training techniques comprising self-supervised learning~\cite{caron2021emerging,jing2020self}, contrastive learning~\cite{grill2020bootstrap,he2020momentum}, language-vision pre-training~\cite{sharma2018conceptual,radford2021learning}, etc. Recently, SAM~\cite{kirillov2023segany} pre-trained on over 11M images stands out as a generalist foundation model to image segmentation and shows powerful zero-shot capabilities of segmenting anything in the wild in an interactive and promptable manner. One of the concurrent works, SEEM~\cite{zou2023segment}, presents a more universal prompting scheme to support semantic-aware open-set segmentation. SegGPT~\cite{wang2023seggpt} further pursues board in-context segmentation tasks in images or videoes.


\noindent \textbf{Parameter-efficient model fine-tuning.}
As the widespread use of foundation models, the topic of parameter-efficient fine-tuning has attracted lots of attention. Existing efficient-tuning methods can be classified into three categories~\cite{ding2023parameter}. Addition-based methods insert lightweight adapters~\cite{houlsby2019parameter,pan2022st,wang2023med} or prompts~\cite{liu2023pre,jia2022visual} into the original model and only tune these parameters. Specification-based methods~\cite{zaken2021bitfit,guo2020parameter} select a small proportion of the original parameters to tune. Rreparameterization-based methods~\cite{hu2021lora} use low-rank matrices to approximate the parameter updates. Recently, there are a few works adapted pre-trained image models to video understanding~\cite{yang2023aim,pan2022st} or volumetric segmentation~\cite{wang2023med}. However, these methods interpret the additional dimension as a "word group", and use the special modules to aggregate the information on that dimension. We consider all three dimensions are isotropic and directly adapt the trained transformer block to catch 3D patterns.

\noindent \textbf{Tumor segmentation in medical imaging.} Tumor segmentation is one of the most common yet challenging tasks in computer-aided medical image analysis.
Recent achievements in deep neural networks have contributed significantly to performance improvement on applications for different anatomical regions, such as liver~\cite{chlebus2018automatic, bilic2023liver}, kidney~\cite{heller2020state}, pancreas~\cite{li2022generalizable} and colon~\cite{pacal2020comprehensive}.
However, precise tumor segmentation is still challenging even for state-of-the-art segmentation networks such as nnU-Net~\cite{isensee2021nnu}, UNETER++~\cite{shaker2022unetr++} and 3D UX-Net~\cite{lee20223d}, because tumors usually have notable properties of small size, irregular shape, low contrast, and ambiguous boundaries. Unsurprisingly, in recently reported SAM applications on medical images, SAM obtained much worse and unstable results on tumor segmentation tasks compared with other anatomical structures such as 3D organs.
Therefore, in this paper, we will focus on evaluating our proposed adapter for tumor segmentation scenarios, in order to address the most significant weakness of the original SAM.

\section{Methods}
\begin{figure}[!tp]
    \centering
    \includegraphics[width=1\linewidth]{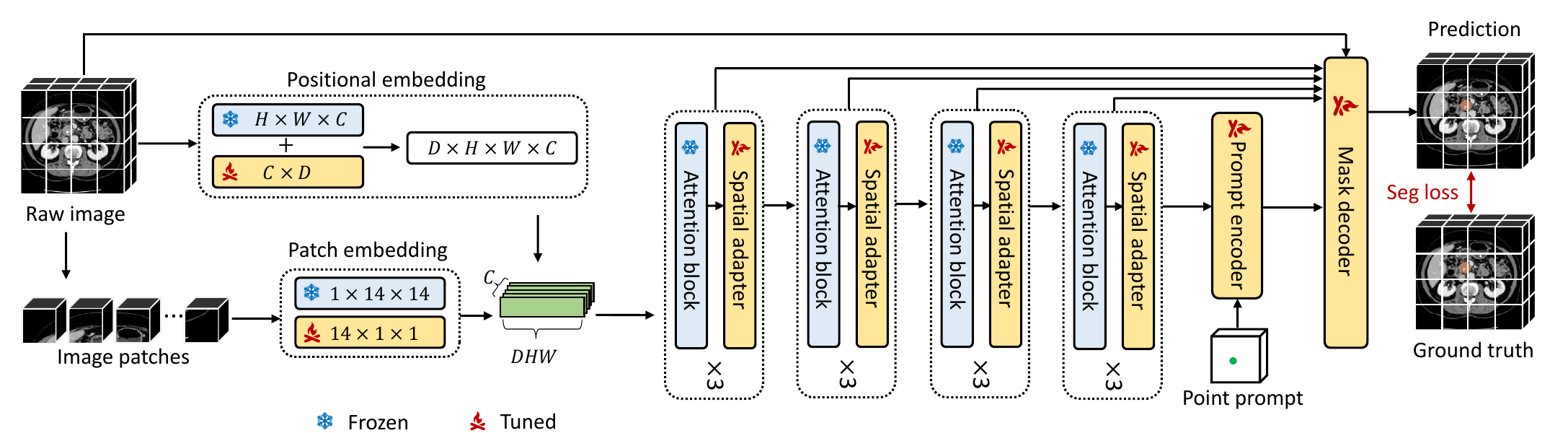}
    \caption{Overview of our proposed method for 3DSAM-adapter. The original ViT is modified to support volumetric inputs. The prompt encoder is redesigned to support 3D point prompt, and the mask decoder is updated to 3D CNN with multi-layer aggregation to generate 3D segmentation.}
    \label{fig:backbone}
\end{figure}

In this section, we will introduce how we adapt the original SAM architecture for volumetric medical image segmentation. Fig.~\ref{fig:backbone} presents the overview of our method.  We first give a brief overview of the SAM, then we explain the technical details for adapting the image encoder, prompt encoder, and mask decoder, respectively.

\subsection{Overview of SAM}
The SAM~\cite{kirillov2023segany} is a large promptable segmentation model with impressive performance and generalization ability on segmenting daily objects. The model consists of three components, i.e., image encoder, prompt encoder, and mask decoder. The image encoder utilizes the structure of the Vision Transformer (ViT)~\cite{dosovitskiy2020image} to transform the original images into image embeddings. The prompt encoder encodes prompts (points, box, etc.) into embedding representations, designed to be lightweight by summating a frozen positional encoding and a learnable embedding for each prompt type. The mask decoder comprises a prompt self-attention block and bidirectional cross-attention blocks (prompt-to-image attention and vice-versa). After conducting attention blocks, the feature map is up-sampled and transformed into segmentation masks by MLP. However, the original structure is designed for 2D natural image segmentation. When transferring to volumetric images, it has to make predictions in a slice-wise manner, which fails to capture the inter-slice spatial information. The model also exhibits performance degradation when being tested on medical images due to the domain gap between natural and medical images. Therefore, task-specific adaptation and fine-tuning are required.

\subsection{Adapting Image Encoder for Volumetric Inputs}
The original SAM is based on 2D ViT, which excels at catching global patterns for natural 2D images. However, many widely adopted medical imaging modalities, such as CT and MRI, are 3D volumes. 3D information is vital for applications such as organ segmentation and tumor quantification since their representative patterns need to be captured from a 3D perspective. Purely relying on 2D views can result in low accuracy due to ambiguous boundaries and non-standard scanning pose.  

Existing methods adapting the 2D pre-trained models for 3D applications usually process the images in a slice-wise manner and then use an additional spatial adaptor or temporal modules to fuse the 2D information~\cite{yang2023aim}. The major parts of the backbone, such as the transformer blocks are still built in 2D. This can work well on video-related tasks but is a suboptimal solution for medical image analysis, as volumetric medical images are isotropic in terms of spatial resolutions and inherent 3D information. It would be problematic to process the third depth dimension differently from the width and height.

To this end, we consider our adaptation method based on two criteria: 1) enabling the model to learn 3D spatial patterns directly, and 2) inheriting most of the parameters from the pre-trained model, and forging incremental parameters of small size and easy-to-tune. Of course, the devil is in the details, as illustrated in Fig.~\ref{fig:backbone}. The original SAM is based on the transformer, which comprises multiple attention blocks and thereby supports inputs of variant token size naturally. Meanwhile, the volumetric medical images are usually isotropic, and the spatial relationship among pixels would be very similar to that of the 2D case. Therefore, we hypothesize the network trained to learn 2D spatial features can be easily adapted to capture 3D patterns as well. The only things remaining are how to initialize the tokens for 3D patches and how to inform a model of the new positional information in a parameter-efficient way. Specifically, we carefully modify each module of the network as follows:

\begin{itemize}
\item \textbf{Patch embedding.} 
We take advantage of the combination of $1\times14\times14$ and $14\times1\times1$ 3D convolutions as an approximation of the $14\times14\times14$ convolution. We initialize the $1\times14\times14$ convolution with the weight of the pre-trained 2D convolution and keep it frozen during the fine-tuning phase. For the newly introduced $14\times1\times1$ 3D convolution, depth-wise convolution is used to further reduce the number of tunable parameters. 
\item \textbf{Positional encoding.} The pre-trained ViT contains a lookup table of size $c \times H\times W$ with the positional encoding. We additionally initialize a tunable lookup table of size $c \times D$ with zeros. The positional encoding of 3D point $(d, h, w)$ can be the summation of embedding in the frozen lookup table with $(h,w)$ and embedding in the tunable lookup table with $(d)$.
\item \textbf{Attention block.} Attention blocks can be directly modified to fit 3D features. For 2D inputs, the size of the queries is $[B, HW, c]$, which can be easily adapted to be $[B, DHW, c]$ for 3D ones with all the pre-trained weights inherited. We use similar sliding-window mechanisms as the Swin Transformer~\cite{liu2021swin} to reduce the memory cost caused by dimension raising.
\item \textbf{Bottleneck.} As convolution layers are usually easier to optimize than transformers~\cite{xiao2021early}, we replace all the 2D convolutions in the bottleneck with 3D ones and train them from scratch.
\end{itemize}

\begin{wrapfigure}{r}{4cm}
    \centering
    \includegraphics[width=0.85\linewidth]{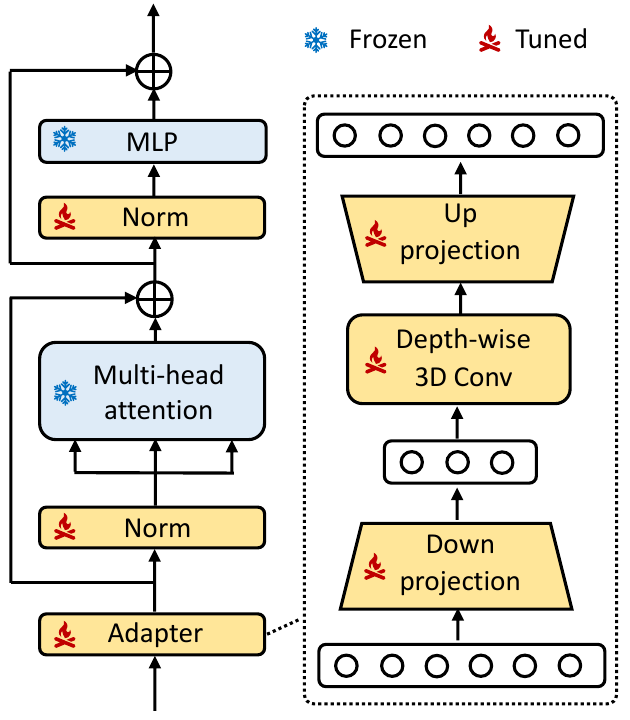}
    \caption{Spatial adapter.}
    \label{fig:adapter}
\end{wrapfigure}
With the above modification, we can elegantly upgrade the 2D ViT to 3D ViT while keeping most of the parameters reusable. Fully fine-tuning the 3D ViT can be memory-intensive. To address this issue, we propose to leverage the lightweight adapter~\cite{houlsby2019parameter} for efficient fine-tuning. The firstly-proposed adapter is composed of a down-projection linear layer and an up-projection linear layer, which can be represented as $\text{Apater}(\textbf{X}) = \textbf{X}+\sigma(\textbf{X}W_{down})W_{up},$ where  $\textbf{X} \in \mathbb{R}^{N\times c}$ is the original feature representation, $W_{down} \in \mathbb{R}^{c\times m}$ and $W_{up} \in \mathbb{R}^{m\times c}$ indicate the down-projection layer and up-projection layer, respectively, and $\sigma(\cdot)$ is the activation function. As illustrated in Fig.~\ref{fig:adapter}, we append a depth-wise 3D convolution after the down-projection layer so the adapter can better leverage 3D spatial information.

During the training phase, we only tune the parameters of convolutions, spatial adapters, and normalization layers, while keeping all other parameters frozen. The frozen scheme makes the training process memory-efficient. Fine-tuning the adapter and normalization layers can help the model narrow the domain gap between natural images and medical images.
\subsection{Prompt Encoding by Visual Sampler}
\begin{wrapfigure}{r}{4cm}
    \centering
    \includegraphics[width=1.0\linewidth]{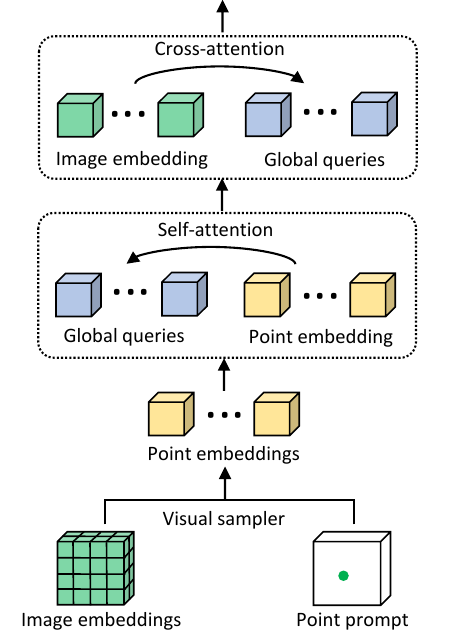}
    \caption{Structure of our prompt encoder based on visual sampler and global queries cross-attention.}
    \label{fig:prompt_encoder}
\end{wrapfigure}
The original SAM leverages positional embedding to represent the prompt. The same way of positional embedding based on Fourier features~\cite{tancik2020fourier} is applied to both the prompt and the image so that the prompt and the image embedding corresponding to the same position can have the same positional encoding. The prompt embedding is then cross-attention with the image embedding and thereby transforms from pure positional features to semantic features. This cross-attention works well for 2D cases but may cause over-smoothing issues~\cite{wang2022anti} when applying to 3D feature maps. Raising to 3D can cause a catastrophic rising in token numbers so that the probability will tend to be a uniform distribution, which makes prompt embedding hard to sufficiently extract semantic information. 

Another issue for medical images is that offering accurate prompt guidance requires lots of professional knowledge, as many background points are actually pretty similar to the foreground ones. The model can easily fail if the prompts are inaccurately given. It is desirable if the system can have higher intelligence, which has a higher tolerance for noisy prompts, thereby supporting non-expert users or even using predicted masks of a coarse segmentation algorithm as prompts.

The third issue is the form of the prompt can be limited for 3D cases, as the bounding boxes can be difficult to draw for volumetric images. A desired method would work even with one point per volume as the prompt, which can be a more challenging setting. Meanwhile, other prompts widely used in the medical domain (e.g., scrawlings) can be transformed into points through sampling. 

To this end and following~\cite{zou2023segment}, we propose to use a visual sampler instead of positional encoding to represent the prompt. The whole process is illustrated in Fig.~\ref{fig:prompt_encoder}. Given the coordinates of the points, we directly interpolate from the feature map to fetch the embeddings, thereby guaranteeing the prompts share the same semantic features with the image embeddings. We initialize a few tokens as global queries and then apply self-attention among the global queries and these points embeddings. After that, we apply cross-attention from image embeddings to these global queries only. As the number of points prompt and global queries are all quite small, this can alleviate the over-smoothing issues.  Besides, this can bring higher tolerance towards noisy points as the global queries can serve as prototypes and only point embeddings with specific features will have high similarities. 

We also include pure background images with false positive prompts during the training phase. At each interaction, if there is no foreground pixel, we randomly sample 10 points from the background as the prompts. Otherwise, we randomly sample about 40 points from the foreground. This training strategy can enhance the model's robustness towards noisy prompt, and the model performs well even if the user gives only one point during the inference phase. 

Note that we discard the original interactive training and inference mode in SAM, where points are added progressively with the next point selected in the misclassified area of the masked predicted from the previous step. It is not as convenient as in the 2D scenery to detect the misclassified area in volumetric segmentation, where the users may need to check every slice. Therefore, we apply a simpler scheme for training and inference, where one or a few point prompts are given all at once.

\subsection{Lightweight Mask Decoder}


The mask decoder of SAM is designed to be lightweight, with stacks of convolution layers. We replace all the 2D convolutions with 3D convolutions to directly generate 3D masks. The initial decoder is designed without any progressive upsampling or skip connection. This works well for nature image where the size of the objects are usually large and the boundary is clear. However, for volumetric medical image segmentation, it is widely acknowledged that U-shape networks with skip connections at multiple levels are critical~\cite{isensee2021nnu}, as the objects in medical images are usually tiny and have ambiguous boundaries, which requires details with higher resolution to be better discriminated. 

\begin{wrapfigure}{r}{6.5cm}
    \centering
    \includegraphics[width=1.0\linewidth]{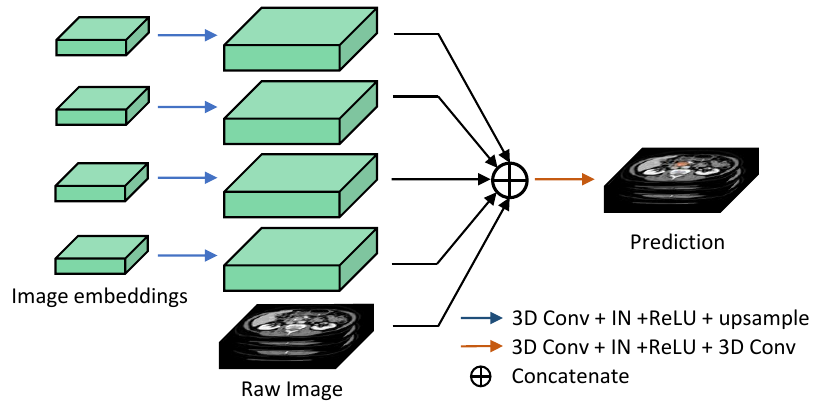}
    \caption{Structure of our lightweight mask decoder with multi-layer aggregation.}
    \label{fig:mla}
\end{wrapfigure}

To alleviate this issue and meanwhile maintain the lightweight property, we utilize a multi-layer aggregation mechanism~\cite{zheng2021rethinking} in our decoder, where the intermediate output of the encoder is concatenated together to produce a mask feature map while the whole structure remains lightweight. To better leverage the information from the original resolution, after upsampling the mask feature map to the original resolution, we concatenate it with the original image and use another 3D convolution to fuse the information and generate the final mask.

We remove multi-masks generation and ambiguity awareness of the original SAM, as our goal is to fine-tune the SAM into a specific downstream task. The backbone of the mask decoder is lightweight and mainly comprises 3D convolutional layers which are optimization friendly, so we train all the parameters from scratch.



Overall, we have introduced a holistic scheme to adapt SAM for medical image segmentation. The adaptation of the image encoder focuses on how to efficiently take advantage of the pre-trained parameters to extract 3D spatial features, while the modification of the prompt encoder and mask decoder mainly resolves the computational issues raised by dimension lifting and also facilitates the model to better align with domain-specific demands. Combining both transforms SAM into a potentially powerful tool for volumetric medical image segmentation.

\section{Experiments}
\subsection{Setup}
\noindent\textbf{Datasets.} We make our experiments focus on tumor segmentation, as this is the reported most challenging task for the original SAM when being applied to medical images. To this end, we employ four public datasets for volumetric tumor segmentation, including: 1) Kidney Tumor Segmentation challenge 2021 dataset (\kits)~\cite{heller2020state}, 2) Pancreas Tumor Segmentation task of 2018 MICCAI Medical Segmentation Decathlon challenge dataset (\msd)~\cite{antonelli2022medical},  3) Liver Tumor Segmentation Challenge 2017 dataset (LiTS17)~\cite{bilic2023liver}, and 4) Colon Cancer Primaries Segmentation task of 2018 MICCAI Medical Segmentation Decathlon challenge dataset (MSD-Colon)~\cite{antonelli2022medical}. The public datasets contain 300, 281, 118 and 126 abdominal CT scans respectively. The original dataset (except for the MSD-Colon) includes both organ and tumor segmentation labels while we are only using tumor labels for training and testing. The datasets are randomly split into 70\%, 10\%, and 20\% for training, validation, and testing. 
More details can be found in the released code and the descriptions in appendix.

\noindent\textbf{Implementation Details.} We implement our method and benchmark baselines in PyTorch and MONAI. We are using SAM-B for all the experiments and comparisons, which utilize ViT-B as the backbone of the image encoder. The model is trained with batch size 1 on an NVIDIA A40 GPU, using AdamW~\cite{loshchilov2017decoupled} optimizer with a linear scheduler for 200 epochs. We set the initial learning rate of $1e{-4}$, momentum of 0.9 and weight decay of $1e{-5}$. We preprocess the data to have isotropic spacings of $1mm$. For data augmentation, we perform random rotation, flip, zoom, and shift intensity. We also randomly sample foreground/background patches at a ratio of 1:1 during training. The complete training details are available in the appendix. Overall, we evaluate the performance of our method via comparisons with current SOTA volumetric segmentation and fine-tuning approaches. The Dice coefficient (Dice) and Normalized Surface Dice (NSD) are used as evaluation metrics.

\subsection{Comparison with State-of-the-Arts}

\begin{figure}[!tp]
    \centering
    \includegraphics[width=1\linewidth]{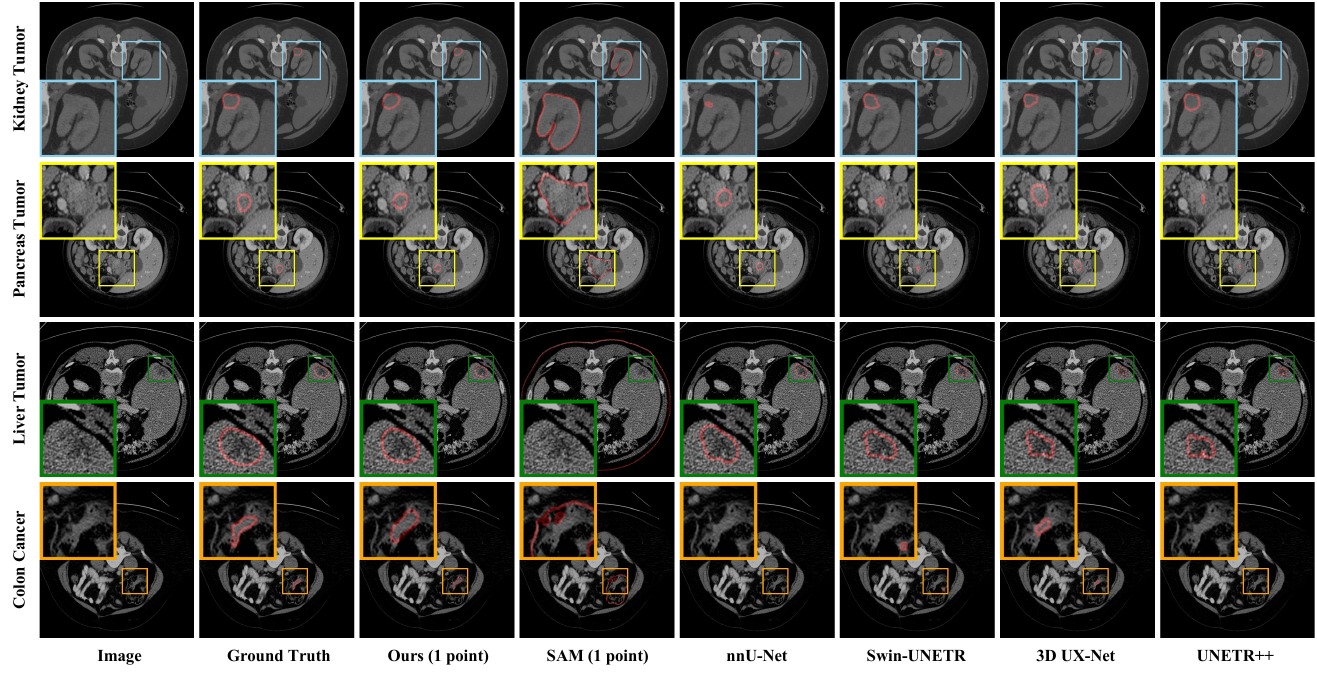}
    \caption{Qualitative visualizations of the proposed method and baseline approaches on kidney tumor, pancreas tumor, liver tumor and colon cancer segmentation tasks.}
    \label{fig:vis}
\end{figure}

\begin{table}[t]
\footnotesize
\centering
\renewcommand\arraystretch{1}
    \caption{Comparison with classical medical image segmentation methods on four tumor segmentation datasets. Well-acknowledged and latest state-of-the-art domain segmentation methods are compared with. Evaluation metrics on Dice score and normalized surface dice (NSD) are reported.}
\begin{center}
\label{tab:sota}
\resizebox{1\textwidth}{!}{
\begin{tabular}{l|cc|cc|cc|cc|c}
\toprule
\multirow{2}{*}{Methods\rule{0pt}{2.4ex}}&\multicolumn{2}{c|}{Kidney Tumor}&\multicolumn{2}{c|}{Pancreas Tumor} & \multicolumn{2}{c|}{Liver Tumor} & \multicolumn{2}{c|}{Colon Cancer} &\multirow{2}{*}{\#Tuned Params\rule{0pt}{2.4ex}}\\ [0.2ex]
 \cline{2-9}
  &Dice~$\uparrow$ &NSD~$\uparrow$&Dice~$\uparrow$&NSD~$\uparrow$&Dice~$\uparrow$&NSD~$\uparrow$&Dice~$\uparrow$&NSD~$\uparrow$ \rule{0pt}{2.4ex}\\
\midrule
nnU-Net (Nat. Methods 2021)~\cite{isensee2021nnu}  &73.07& 77.47 &41.65& 62.54&\textbf{60.10}& \textbf{75.41} &43.91& 52.52& 30.76M \\
TransBTS (MICCAI 2021)~\cite{wang2021transbts}&40.79&37.74& 31.90 & 41.62&34.69&49.47&17.05&21.63&32.33M\\
nnFormer (arXiv 2021)~\cite{zhou2021nnformer}&45.14&42.28&36.53& 53.97&45.54&60.67&24.28&32.19&149.49M\\
 Swin-UNETR (CVPR 2022)~\cite{tang2022self} &65.54&72.04&40.57&60.05&50.26&64.32&35.21&42.94&62.19M\\
 UNETR++ (arXiv 2022)~\cite{shaker2022unetr++}&56.49&60.04&37.25&53.59&37.13&51.99&25.36&30.68&55.70M\\
 3D UX-Net (ICLR 2023)~\cite{lee20223d} &57.59&58.55&34.83&52.56&45.54&60.67&28.50&32.73&53.01M\\
 \midrule
SAM-B (1 pt/slice)~\cite{kirillov2023segany} &36.30&29.86&24.01&26.74&6.71&7.63&28.83&33.63&--\\
Ours (1 pt/volume) &\textbf{73.78}&\textbf{83.86}&\textbf{54.09}&\textbf{76.27}&54.78&69.55&\textbf{48.35}&\textbf{63.65}&25.46M\\
\midrule
SAM-B (3 pts/slice))~\cite{kirillov2023segany} &39.66&34.85&29.80&33.24&7.87&6.76&35.26&39.31&-- \\
Ours (3 pts/volume) &\textbf{74.91}&\textbf{84.35}&\textbf{54.92}&\textbf{77.57}&56.30&70.02&\textbf{49.43}&\textbf{65.02}&25.46M\\
\midrule
SAM-B (10 pts/slice))~\cite{kirillov2023segany} &40.07&34.96&30.55&32.91&8.56&5.97&39.14&42.70&--\\
Ours (10 pts/volume) &\textbf{75.95}&\textbf{84.92}&\textbf{57.47}&\textbf{79.62}&56.61&69.52&\textbf{49.99}&\textbf{65.67}&25.46M\\
\bottomrule
\end{tabular}}
\end{center}
\end{table}

We extensively compare our model with the recent 3D medical image segmentation state-of-the-art, including the most recent Transformer-based methods including Swin-UNETR~\cite{tang2022self}, UNETR++~\cite{shaker2022unetr++}, TransBTS~\cite{wang2021transbts}, and nnFormer~\cite{zhou2021nnformer}, as well as CNN-based methods such as nnU-Net~\cite{isensee2021nnu} and 3D UX-Net~\cite{ji2020uxnet}. As reported in Table~\ref{tab:sota}, we observe that the original SAM~\cite{kirillov2023segany} developed in natural images gets suboptimal performances in domain-specific tumor segmentation tasks, even with as many as ten clicks being prompts for small tumors. With the proposed effective and efficient adaptation, we obtain state-of-the-art accuracy with a single point only in the whole volume, outperforming both the powerful medical image segmentation baselines and nnU-Net~\cite{isensee2021nnu}, which is widely used in volumetric segmentation competitions. Distinct improvements can be specifically observed for pancreas tumors and colon cancers, which are always notorious for ambiguous boundaries, of 12.44\% and 10.11\% in Dice against the prior state-of-the-art. Besides, we observe stable performance rising when feeding more points for prompting. Our model gets Dice of 75.95\%, 57.47\%, 56.61\% and 49.99\% in kidney tumor, pancreas tumor, liver tumor and colon cancer segmentation respectively with 10 points per volume (which is usually affordable in common clinical usage). Relevant evidence is also observed in Fig.~\ref{fig:vis}, where SAM usually fails to distinguish tumor borders and gives false positive predictions, whereas our adaptation demonstrates visually better tumor identification results in most tasks. The performance of liver tumor segmentation is inferior to that of nnU-Net~\cite{isensee2021nnu}, which is reasonable because the liver tumor is comprised of multiple smaller tumors, scattered around the liver. The prompt-based method can only click one of them, leaving the rest unsegmented and therefore resulting in many false negative pixels. Nevertheless, compared with the original SAM, our model still shows significant improvements. 

\subsection{Comparison with Existing Adapters}
We further compare our adaptation strategy with existing parameter-efficient adaptation methods, which includes 2D adaptations such as adapter~\cite{houlsby2019parameter} and Pro-tuning~\cite{nie2022pro}, as well as 2D-3D adapters such as ST-Adapter~\cite{pan2022st} and Med-Tuning~\cite{wang2023med}. For 2D adaptations, we encode the images in a slice-wise manner and then concatenate them to form a 3D feature map before plugging them into the decoder. We also compare with the method of full fine-tuning, which tunes all the parameters of the original SAM. To conduct a fair comparison and be more focused on whether adaptations help the model learn volumetric medical image features,  we remove the prompt encoder and only train the model with an image encoder followed by a lightweight decoder (STER-MLA~\cite{zheng2021rethinking}). So that the model will conduct automatic segmentation without prompt guidance. The performance can well reflect the presentation learning ability of the encoder. All the pre-trained weights are from SAM-B. 

The results are given in Table~\ref{adaptation}, which shows our adaptation strategy outperforms all existing methods with a comparable number of tuned parameters. Our method excels the second-best method by 16.24\% on kidney tumor segmentation Dice, 4.22\% on pancreas tumor segmentation NSD and 12.53\% on Colon Cancer segmentation Dice. It even outperforms many classical segmentation methods with fewer tunable parameters and a vanilla mask decoder. Our method also outperforms full fine-tuning of SAM by 2.08\% $\sim$ 29.80\% with only less than 16.96\% tunable parameters. These results can well substantiate our hypothesis that parameters pre-trained on 2D images can be used to learn 3D spatial features with only minor adaptations. And treating all dimensions equally is actually a better strategy than interpreting the depth dimension as a group in medical image segmentation, especially when the images have similar spacings in all dimensions.
\begin{table}[t]
\footnotesize
\centering
\renewcommand\arraystretch{1}
\caption{Comparison with existing parameter-efficient and full fine-tuning methods. We discard the prompt encoder and only tune the image encoder and mask decoder for fully automatic segmentation.}
\label{adaptation}
\begin{center}
\label{adaptation}
\resizebox{1\textwidth}{!}{
\begin{tabular}{l|cc|cc|cc|cc|c}
\toprule
\multirow{2}{*}{Methods\rule{0pt}{2.4ex}}&\multicolumn{2}{c|}{Kidney Tumor}&\multicolumn{2}{c|}{Pancreas Tumor} &\multicolumn{2}{c|}{Liver Tumor} &\multicolumn{2}{c|}{Colon Cancer} &\multirow{2}{*}{{\makecell[c]{\#Tuned Params \\(image encoder)}}\rule{0pt}{2.4ex}}\\ [0.2ex]
 \cline{2-9}
  &Dice~$\uparrow$ &NSD~$\uparrow$&Dice~$\uparrow$&NSD~$\uparrow$ \rule{0pt}{2.4ex}&Dice~$\uparrow$&NSD~$\uparrow$ \rule{0pt}{2.4ex}&Dice~$\uparrow$&NSD~$\uparrow$ \rule{0pt}{2.4ex}& \\
\midrule
 Full fine-tuning &52.31&50.35&26.49&33.28&45.59&52.53&24.63&\textbf{40.67}&89.67M  \\  
 \midrule
 
 Adapter (ICML 2019)~\cite{houlsby2019parameter}&46.99&43.76&20.28&30.81&42.17&57.52&22.55&38.10&7.61M\\
 Pro-tuning (arXiv 2022)~\cite{nie2022pro}&50.73&50.81&18.93&30.45&47.33&55.61&21.24&25.10&7.17M\\
 ST-Adapter (NeurIPS 2022)~\cite{pan2022st}&47.30&45.61&\textbf{30.27}&43.53&51.93&59.93&28.41&34.60&7.15M\\
 Med-tuning (arXiv 2023)~\cite{wang2023med}&44.73&40.28&22.87&30.02&\textbf{52.06}&\textbf{68.44}&21.08&37.78&11.10M\\
 \midrule
 Ours&\textbf{61.60}&\textbf{70.40}&30.20&\textbf{45.37}&49.26&59.48&\textbf{31.97}&\textbf{40.67}&15.21M\\
\bottomrule
\end{tabular}}
\end{center}
\end{table}

\subsection{Ablation Studies}
We also conduct comprehensive ablation studies regarding the design of the prompt encoder and mask decoder. 
Without loss of generality, the experiments are conducted on the KiTS 21 dataset and for each trial, we only give one point as a prompt while each result is based on 10 random trials.

\begin{figure}[!tp]
    \centering
    \includegraphics[width=1\linewidth]{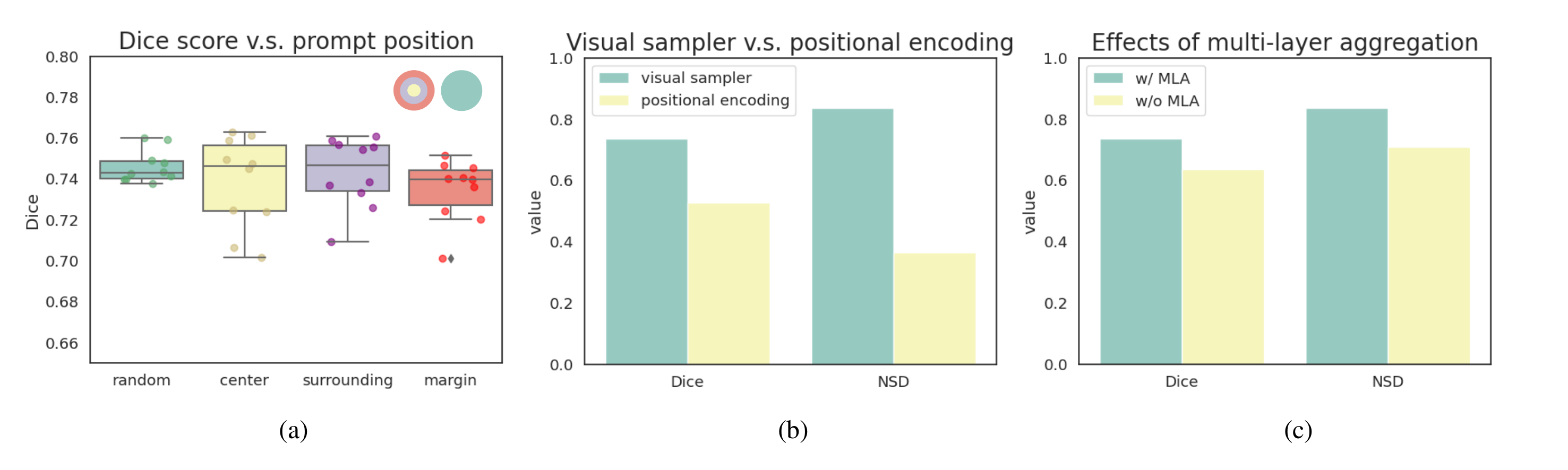}
    \caption{Results on our ablation studies. (a) Performance v.s prompt. (b) Visual sampler v.s. positional encoding. (c) Effects of multi-layer aggregation.} 
    \label{fig:ablation}
\end{figure}

\noindent \textbf{Positional encoding v.s. visual sampler.}
In order to prove visual sampler is actually a better way for prompt encoding compared with the original positional encoding method used in SAM in terms of volumetric segmentation, we fix the image encoder and mask decoder while only changing the prompt encoder. The results are shown in Fig.~\ref{fig:ablation}. Our proposed visual sampler algorithm outperforms the original positional encoding by 40.00\% for Dice. This is align with our claim that visual sampler work well when the size of tokens is large as it does not have over-smoothing issues.
\begin{wrapfigure}{r}{5cm}
    \centering
    \includegraphics[width=1.0\linewidth]{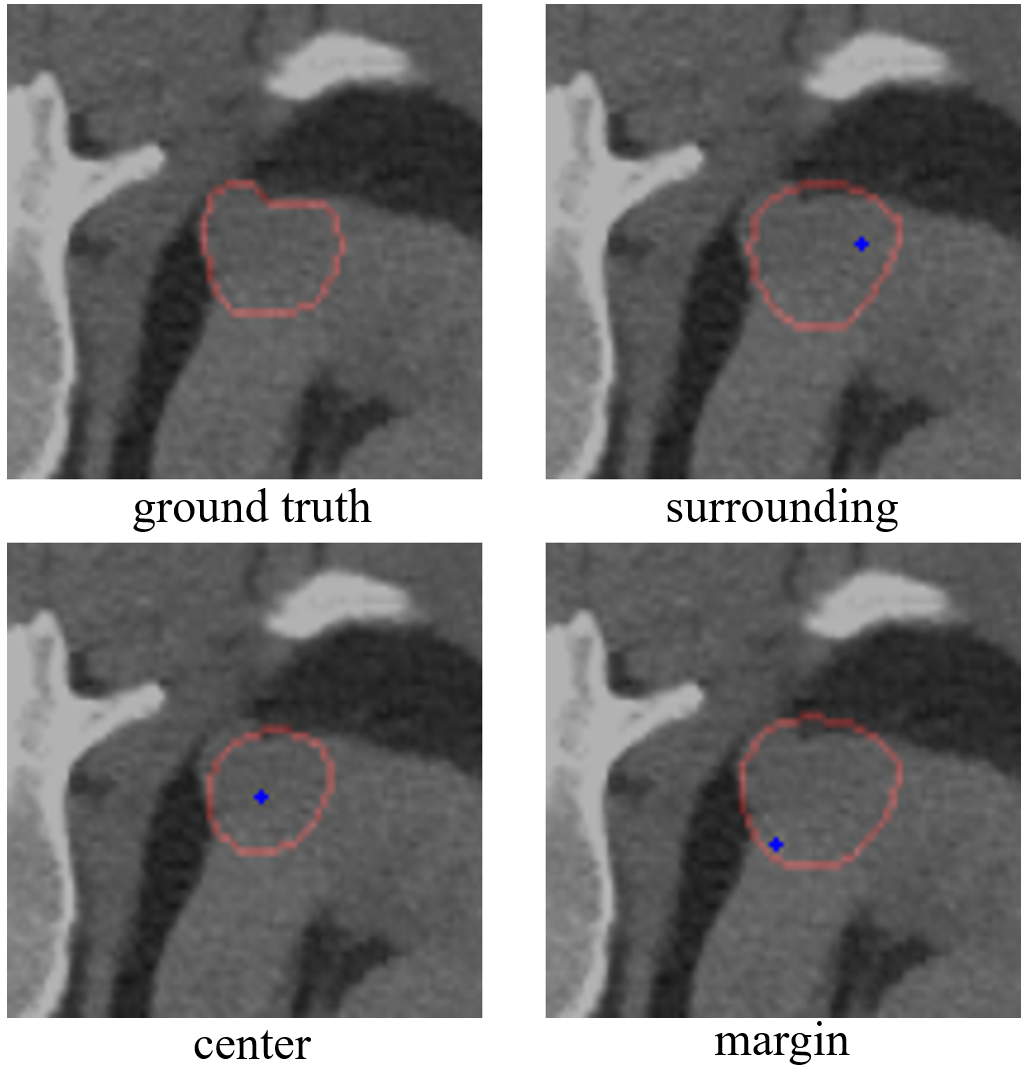}
    \caption{Segmentation results in terms of different prompt location. Blue marker denotes the prompt.}
    \label{fig:location}
\end{wrapfigure}

\noindent \textbf{Position of the point prompts.}
 We analyze how the model performance is different when the point prompt is given at the center of the objects, at its adjacent regions, or at the margin. We divide the ground truth tumor mask into three regions, the center region, the surrounding region, and the marginal region, as illustrated in Fig.~\ref{fig:location}. For cases with multiple tumors, we are sampling from the one with the largest size. From the results in Fig.~\ref{fig:ablation}, we find that the model is not very sensitive to the position of the point prompts. Giving prompt at different positions yield almost the same results.

\noindent \textbf{Effects of multi-layer aggregation.} To verify  our modification of the mask decoder actually brings performance gain, we compare it with the one without multi-layer aggregation. The result Fig.~\ref{fig:ablation} shows the model with multi-layer aggregation has 15.75\% higher Dice than its counterpart. This also aligns with common practice in medical imaging segmentation as local texture features are very important, especially for tumor segmentation which usually has tiny shapes and low contrast.

\noindent \textbf{Effects of deep prompts.} Since it is beneficial to include features from multiple levels of the image encoder, a natural idea comes that incorporating prompts at multiple levels might also improve performance. To this end, we conduct experiments to plug in the prompt encoder to other feature levels besides the final bottleneck levels. The results are shown in Table~\ref{deepprompt}. However, we find incorporating deep prompts brings no gain or even degenerates the performance. This can be left as a topic for further research to make prompts take effect at multiple feature levels. 
\section{Discussion and Conclusion}
In this work, we propose a comprehensive scheme to adapt SAM from a 2D natural image generalist to a volumetric medical imaging expert, especially for tumor segmentation. Through parameter-efficient fine-tuning, our method significantly improves SAM's performance in the medical domain, making it outperform SOTA, with only a single coarse click as the prompt. Our proposed method can also beat existing adaptation methods for volumetric adaptation. 
\begin{wraptable}{r}{5cm}
\footnotesize
\centering
\renewcommand\arraystretch{1}
\caption{Comparison of models w/o deep prompts.}
\label{deepprompt}
\begin{center}
\begin{tabular}{c|c|c|c|c|c}
\toprule
\multicolumn{4}{c|}{Feature Level}&\multirow{2}{*}{Dice~$\uparrow$\rule{0pt}{2.6ex}}&\multirow{2}{*}{NSD~$\uparrow$\rule{0pt}{2.6ex}}\\ [0.2ex] 
 \cline{1-4}
 1&2&3&4&& \rule{0pt}{2.6ex}\\
\midrule
\checkmark&\checkmark&\checkmark&\checkmark&64.61&71.33\\
\checkmark&&&\checkmark&66.82&74.87\\
&\checkmark&&\checkmark&69.17&77.49\\
&&\checkmark&\checkmark&65.39&72.56\\
&&&\checkmark&\textbf{73.78}&\textbf{83.86}\\
\bottomrule
\end{tabular}
\end{center}
\end{wraptable}
\noindent \textbf{Limitations and future work.}
In the experiment, our reported metric is lower than that on official leaderboards. This is because we are not taking advantage of the organ segmentation label for multi-label prediction. Although multi-task segmentation brings superior performance gain, pure tumor segmentation is more clinically meaningful as it can bypass the intensive work of organ labeling. One observation is that although many transformer-based methods can outperform nnU-Net for multi-class segmentation, for pure tumor segmentation, the general trend is that CNN-based methods have better performance and are easier to train. This maybe because the size of the tumor is very small, and tumor detection relies more on local texture information. So the global information, which is the strength of the transformer, is not very useful. This poses great obstacles for our work as SAM is based on ViT and lots of detailed texture information can be lost during the first downsampling operation. Our future direction may entail how to adapt the architecture to recover these texture details so that the performance can achieve SOTA even in a fully automatic manner.

\noindent \textbf{Social impact.} Our work transforms SAM into a powerful medical image segmentation tool, which has the potential to be put into real clinic use. It can significantly improve the segmentation accuracy and therefore benefits related diagnostic tasks. Additionally, our pioneering work can serve as a more general guideline on how to adapt foundation models for domain-specific downstream tasks, which can help the ever-increasing foundation models play more important roles in all industries.

{
\small
\bibliography{ref}
\bibliographystyle{unsrt}
}
\newpage
\appendix
\section{Appendix}

In this supplementary material, we provide additional
details about the datasets and the implementation that are not
included in the main paper due to the space limit. 

\subsection{Datasets and Implementation Details}
We summarize the details and data preprocessing hyper-parameters of public datasets used in our experiments in Table~\ref{tab:data}. We apply hierarchical steps for data preprocessing: 1) resampling anisotropic samples at the same target spacing. For anisotropic images, resampling is performed at each plane independently~\cite{isensee2021nnu}. 2) intensity clipping is applied to enhance the contrast of target tissues, where the 0.5 and 99.5 percentiles of the foreground voxels intensity are used for clipping. 3) Intensity normalization is performed with the mean and s.d. of global foreground voxels. We then randomly crop sub-volumes at the ratio of $\text{foreground}:\text{background} = 1:1$ for training. For data augmentation, we perform random rotation, flip, zoom, and intensity shifting. We train our model using AdamW optimizer for 200 epochs with the base learning rate $1e^{-4}$. We conduct all the experiments on NVIDIA A40 GPUs. One epoch of training our model takes around 16 minutes for Kidney Tumor, 13 minutes for Pancreas Tumor, 6 minutes for Liver Tumor, and 7 minutes for Colon. For inference of automatic segmentation methods, we are using sliding window inference method. For inference with point prompt, we calculate the center of the prompts and then crop a small patch surrounding this center, which has the same size as that for training, and do inference on the particular patch. Complete hyper-parameters used in our implementation are summarized in Table~\ref{tab:augmentation}.
\begin{table}[h!]
\footnotesize
\centering
\renewcommand\arraystretch{1}
    \caption{Hyper-parameters of training on public datasets.}
\begin{center}
\label{tab:augmentation}
\noindent\begin{tabular}{l|c}
\toprule

Training Epochs & 200\\
\#Cropped Patches & 1\\
Batch Size & 1\\
AdamW $\beta$ & $(0.9, 0.999)$ \\
AdamW $\lambda$ & $1e^{-5}$ \\
Base Learning Rate & $1e^{-4}$\\
Learning Rate Scheduler & Linear \\ 
End Factor & 0.1 \\
\midrule
Data Augmentation & Random Zoom, Rotation $90^\circ$,  Flip, Intensity Shift \\
Cropped Foreground:Background Ratio & 1:1\\
Zoom Scale \& Probability & [0.9, 1.1], 0.3\\
Rotation $90^\circ$ Probability & 0.5 \\
Flip Probability & 0.5 \\
Intensity Offset & 0.1 \\
\midrule
NSD Tolerance & 5\\
Sliding Window Inference Overlap & 0.7\\
Sliding Window Inference Mode & Constant \\

\bottomrule
\end{tabular}
\end{center}
\end{table}
\begin{table}[h!]
\footnotesize
\centering
\renewcommand\arraystretch{1}
    \caption{Detailed overview of public datasets.}
\begin{center}
\label{tab:data}
\resizebox{1\textwidth}{!}{
\noindent\begin{tabular}{l|cccc}
\toprule
Datasets & Kidney Tumor\cite{heller2020state} & Pancreas Tumor\cite{antonelli2022medical} & Liver Tumor\cite{bilic2023liver} & Colon Cancer\cite{antonelli2022medical}\\ 
\midrule
Imaging Modality & \multicolumn{4}{c}{Multi-Contrast CT} \\
Anatomical Region & \multicolumn{4}{c}{Abdomen} \\
Sample Size & 300 & 281 & 118 & 126\\
Dimensions (Minimum) & $512 \times 512\times 29$ & $512 \times 512\times 37$ & $512 \times 512\times 74$ & $512 \times 512\times 37$\\
Dimensions (Maximum) & $512 \times 512\times 1059$ & $512 \times 512\times 751$ & $512 \times 512\times 987$ & $512 \times 512\times 729$\\
Spacing (Minimum) & $0.44\times0.44\times0.50$ & $0.61\times0.61\times0.70$ & $0.56\times0.56\times0.70$ & $0.54\times0.54\times 1.25$\\
Spacing (Maximum) & $1.04\times1.04\times5.00$ & $0.98\times0.98\times7.50$ & $1.00\times1.00\times5.00$ & $0.98\times0.98\times 7.50$\\
Target Anatomical Label & Kidney Tumor & Pancreas Tumor & Liver Tumor & Colon\\
\midrule
Intensity Clipping Range & [-52, 247] & [-39, 204] & [-17, 201] & [-57, 175]\\
Resampled Spacing & \multicolumn{4}{c}{(1, 1, 1)}\\
Global Foreground Voxel Intensity Mean & 59.54 & 68.45 & 99.40 & 65.18\\
Global Foreground Voxel Intensity S.D. & 55.46 & 63.42 & 39.39 & 32.65\\
Patch Size & $256\times256\times256$ & $128\times128\times128$ & $128\times128\times128$ & $256\times256\times256$\\

\bottomrule
\end{tabular}}
\end{center}
\end{table}

\end{document}